\documentclass[letterpaper, 10 pt, conference]{ieeeconf}
\IEEEoverridecommandlockouts
\usepackage{amsmath,amsfonts}
\usepackage{array}
\usepackage[caption=false,font=footnotesize,labelfont=rm,textfont=rm]{subfig}
\usepackage{textcomp}
\usepackage{stfloats}
\usepackage{url}
\usepackage{verbatim}
\usepackage{graphicx}
\usepackage{cite}
\usepackage{booktabs}
\usepackage{hyperref}
\usepackage[ruled]{algorithm2e}
\usepackage{hyperref}

\begin{document}

\author{Yuhong Cao$^{1}$, Rui Zhao$^{1}$, Yizhuo Wang$^{1}$, Bairan Xiang$^{1}$, Guillaume Sartoretti$^{1}$
\thanks{$^{1}$ Author is with Department of Mechanical Engineering, College of Design and Engineering, National University of Singapore.
        {\tt\small caoyuhong@u.nus.edu}}}

\title{Deep Reinforcement Learning-based Large-scale Robot Exploration}

\maketitle

\begin{abstract}
In this work, we propose a deep reinforcement learning (DRL) based reactive planner to solve large-scale Lidar-based autonomous robot exploration problems in 2D action space. Our DRL-based planner allows the agent to reactively plan its exploration path by making implicit predictions about unknown areas, based on a learned estimation of the underlying transition model of the environment. To this end, our approach relies on learned attention mechanisms for their powerful ability to capture long-term dependencies at different spatial scales to reason about the robot's entire belief over known areas. Our approach relies on ground truth information (i.e., privileged learning) to guide the environment estimation during training, as well as on a graph rarefaction algorithm, which allows models trained in small-scale environments to scale to large-scale ones. Simulation results show that our model exhibits better exploration efficiency ($12\%$ in path length, $6\%$ in makespan) and lower planning time ($60\%$) than the state-of-the-art planners in a $130m\times 100m$ benchmark scenario. We also validate our learned model on hardware.
\end{abstract}

\section{Introduction}

Autonomous exploration focuses on finding the shortest total exploration path to map an unknown environment (i.e., classify it into free and occupied areas). In this work, we consider autonomous exploration based on omnidirectional 3D Lidar with 2D action space (i.e., for a ground robot). There, recent developments of Lidar-based simultaneous localization and mapping (SLAM) can now near-perfectly mitigate state estimation error and yield high-quality maps on hundreds-meter scenarios~\cite{zhang2014loam}. Leveraging these advanced SLAM approaches, current autonomous exploration planners~\cite{selin2019efficient,dang2020graph,cao2021tare,yang2021graph,cao2023representation} are able to assume the localization and mapping are perfect and start tackling complex larger-scale environments towards real-life deployments.
However, optimizing trajectories in large-scale environments is non-trivial, since the exploration task requires real-time replanning as the robot discovers unknown areas and updates its map, with planning horizons in the hundreds of meters.

\begin{figure}
    \centering
    \includegraphics[width=0.4\textwidth]{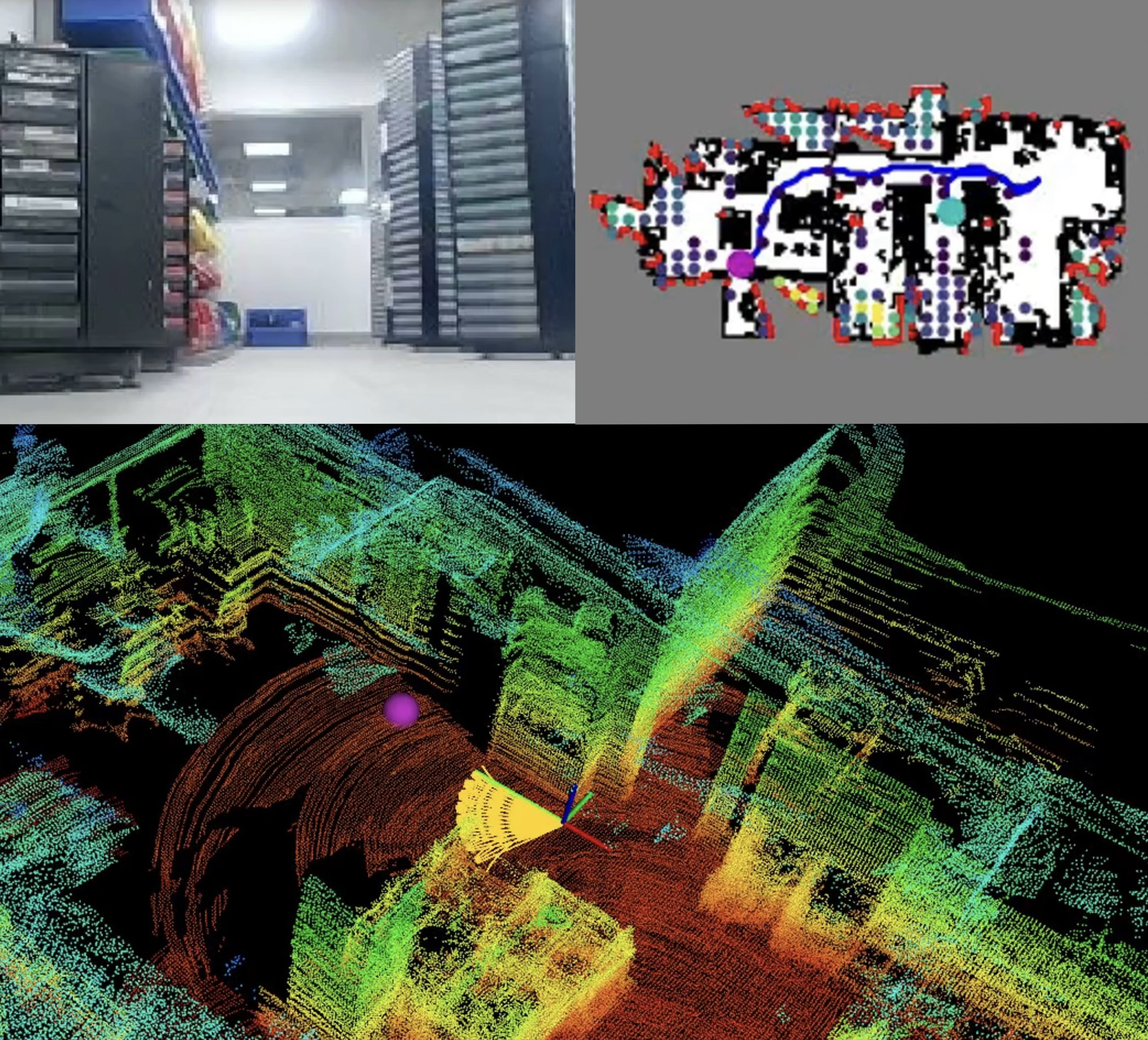}
    \caption{
    \textbf{A mobile ground robot exploring an indoor lab environment using our DRL-based planner.} Top-left: view from the robot's onboard camera (\textit{not used for mapping}). Top-right: agent's current belief (map) of the environment, and view of our neural network's inputs.
    Bottom: 3D point cloud (map) of the environment constructed by the robot. The axes system represents the current position and orientation of the robot, the purple ball the next waypoint output by our planner,}
    \label{fig: exploration example}
    \vspace{-0.5cm}
\end{figure}

Advanced large-scale planners often leverage hierarchical planning to decrease computational complexity while maintaining the fineness of local paths~\cite{selin2019efficient,dang2020graph,cao2021tare,cao2023representation}, as a result, the state-of-the-art planner, TARE~\cite{cao2021tare}, is able to explore environments at the hundred-meters scale while keeping computing times under a second at each planning step.
Nevertheless, these conventional planners share the same limitation: they try to optimize paths solely based on the partial belief (map) over the environment.
However, even optimal long-term paths on the partial map can lead to suboptimal longer-term behaviors as the map is updated with new measurements that may contradict the current plan.
At its core, the exploration problem is a partially observable Markov decision process (POMDP), where the transition model between true states (ground truth of the environment) and the robot belief (map) is hard to predict.
For example, a robot cannot confirm that two close-by, incomplete corridors in its partial map are connected before fully exploring this portion of the environment.
However, we note that humans in the same situations will often try to infer the complete structure of an environment from a partial map (i.e., guess and reconstruct that full corridor) to improve efficiency.
Learning-based approaches such as supervised learning or model-based reinforcement learning may be an option to explicitly predict unknown areas from the agent's partial map, allowing conventional planners to then plan paths over the predicted belief.
However, since the underlying transition model in autonomous exploration is highly stochastic and thus very hard to estimate/learn, directly relying on such a learned prediction may not be a wise choice in practice.

We believe that a simple model-free deep reinforcement learning (DRL) approach can elegantly handle all these challenges, by implicitly estimating the transition model and learning an adaptive policy.
We build upon recent attention-based policy neural network~\cite{cao2022catnipp,cao2023ariadne}, which have shown outstanding abilities at allowing the agent to reason about its entire belief at multiple spatial scales.
In this work, we further propose to train an attention-based policy with the assistance of ground truth knowledge (i.e., privileged learning). It enables the model to draw out more potential of attention mechanism at implicitly predicting unknown areas from the partial robot belief, leading to a more stable training process and better performance after training.
In particular, we train our model under the soft actor-critic (SAC) algorithm~\cite{haarnoja2018soft,christodoulou2019soft}, and let the critic network have access to the ground truth of the environment to get precise long-term evaluations.
We further propose a graph rarefaction algorithm, which enhances the model's capacity to capture longer-term temporal dependencies between areas, thus extending the use of trained policy from small- to large-scale environments.
We compare our DRL-based planner with conventional planners in simplified small-scale environments, where it achieves 11\% better exploration efficiency in terms of path length.
We then test our planner in a high-fidelity 3D Gazebo simulation benchmark of a $130m\times 100m$ indoor office, where it achieves better exploration efficiency (12\% in path length, 6\% in makespan) and significantly lower computing times (60\%) compared to the state-of-the-art exploration planner: TARE~\cite{cao2021tare,cao2023representation}.
Finally, we experimentally validate our planner on a ground robot in an $80m\times 10m$ indoor scenario involving cluttered furniture, highlighting its real-life applicability without any additional training (demonstrated in Figure~\ref{fig: exploration example}).
To the best of our knowledge, our work is the first DRL planner that can effectively explore such large-scale environments.

\section{Related Works}

In this work, we focus on methods applicable to the problem setup where the information in the robot belief is evaluated through frontiers (i.e., boundary between known, free area and unknown area). There are another group of other methods based on the problem setup where the information is evaluated through information theory (e.g., mutual information)~\cite{zhang2020fsmi,asgharivaskasi2022active, asgharivaskasi2023semantic}. However, information theory-based methods require the robot belief to be a probabilistic occupancy map, making them not applicable in the setup of this work and related works discussed below. 

\noindent \textbf{Small-scale exploration planner} Frontier-based exploration planners have been shown to be very efficient in small-scale environments. Yamauchi et al.~\cite{yamauchi1997frontier} first proposed to utilize frontiers to drive exploration, where the \textit{cost} (i.e., the path length from the current position to a frontier) is usually used to rank frontiers, with the lowest-costed frontier being chosen as the next one to be visited by the robot.
More recent works~\cite{holz2010evaluating,kulich2011distance} further considered \textit{utility} (i.e., the size of a frontier, quantifying the amount of information expected to be collected at that frontier) to improve exploration efficiency.
Although the resulting paths are often short-sighted (i.e., optimize for shorter-term objectives, often at the cost of longer-term exploration ones), the above planners remain near-optimal in simple scenarios, where the number of frontiers is relatively low.
In general, as the number of frontiers increases in larger and more cluttered environments, frontier-based planners suffer from such \textit{myopic} decisions. 
Instead of evaluating a single frontier,~\cite{bircher2016receding} relied on a sampling-based strategy in a receding-horizon manner to optimize a long-term trajectory, thus achieving non-myopic decisions and significantly outperforming frontier-based planners in relatively complex (but still small-scale) environments while remaining executable in real-time.

\noindent \textbf{Large-scale exploration planner} Optimizing long-term trajectories online in large-scale exploration problems is challenging since trajectories might be over $100m$ in length, while the gap between individual waypoints should remain below $1m$ to allow for fine detail mapping. Some advanced sampling strategies have been proposed to tackle larger-scale exploration~\cite{duberg2022ufoexplorer, schmid2020efficient}. 
Selin et al.~\cite{selin2019efficient} found that, in large-scale environments, frontiers are sparse at the global scale but dense at the local level, and proposed a hierarchical framework that combined a global frontier-based planner and a local planner based on~\cite{bircher2016receding}.
Following works~\cite{dang2020graph,zhou2021fuel,cao2023representation,cao2021tare} drew on the experience of such hierarchical frameworks to perform path planning at multiple resolutions and further improve the exploration efficiency of planned paths, which significantly outperformed naive frontier- and sampling-based planners~\cite{dang2020graph,cao2023representation,cao2021tare}.
Notably, Cao et al.~\cite{cao2023representation,cao2021tare} proposed to use a TSP-based global planner and a sampling-based local planner with a coverage constraint to trade off the planning fineness of nearby and distant areas, which currently exhibits state-of-the-art performance with 80\% better exploration efficiency over other benchmark planners~\cite{bircher2016receding,mbplanner2020,dang2020graph}.

\noindent \textbf{DRL-based exploration planner} 
While some works~\cite{luperto2021exploration,schmid2022fast} have relied on deep learning to explicitly predict
the layout of the unknown (but only structured indoor) environments to assist conventional approaches, most learning-based approaches leverage deep reinforcement learning
(DRL) to directly and reactively make decisions on movements. DRL-based planners are expected to achieve long-term exploration objectives by training a policy to maximize long-term \textit{returns}.
Most current DRL-based exploration planners~\cite{zhu2018deep,chen2019self,li2019deep,xu2022explore} rely on convolutional neural networks (CNNs) to select a waypoint from a set of locations, based on a visual representation of the agent's belief (e.g., map).
Although DRL intuitively looks like a well-suited method for autonomous exploration, the performance of these DRL-based planners is only on par with naive frontier-based planners in small-scale environments.
On the other hand, Chen et al.~\cite{chen2020autonomous} proposed to utilize graph neural network to handle the localization error during exploration, however their approach can only work in environments without obstacles.
Notably, relying on learned attention over a graph representation of the agent's belief, our previous work ARiADNE~\cite{cao2023ariadne} achieved remarkable improvements over frontier-based planners and one of these CNN-based DRL planners~\cite{chen2019self}.
However, ARiADNE's advantage over TARE is marginal, and like other previous DRL methods, this DRL planner cannot scale to large environments.

\begin{figure*}
    \vspace{+0.1cm}
    \centering
    \includegraphics[width=0.95\textwidth]{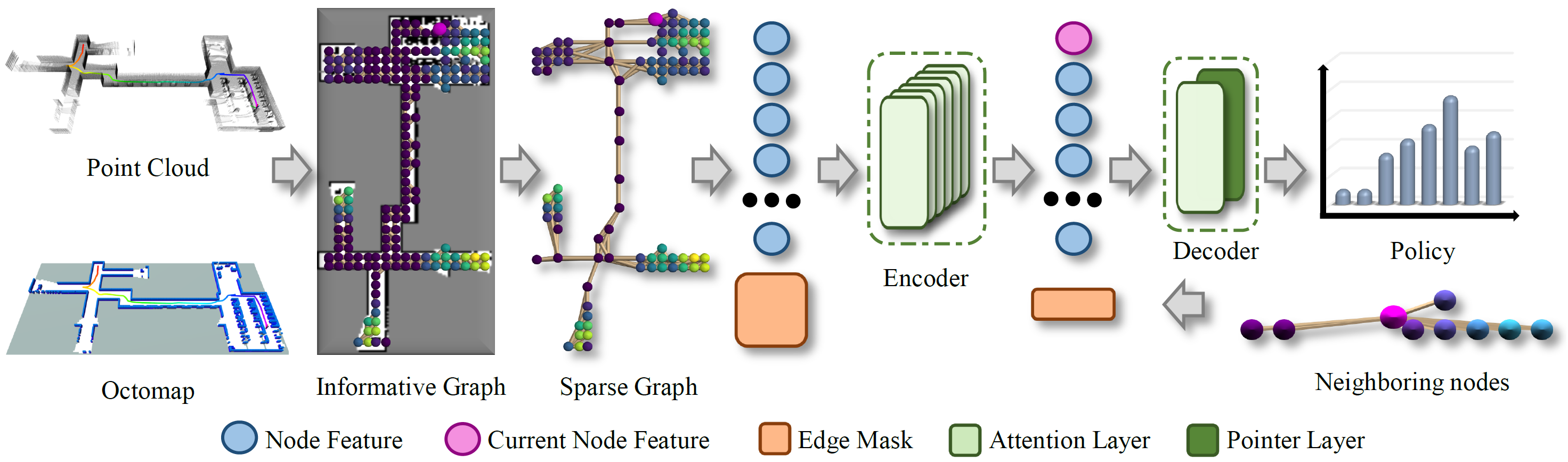}
    \vspace{-0.2cm}
    \caption{\textbf{Our proposed DRL-based planner.} The robot first transforms its current map (point cloud) data to an occupancy grid, and then extracts and rarefies an informative graph from it.
    After that, this graph (node features and an adjacency edge mask) is fed to our attention-based network (consisting of an encoder and a decoder), which finally outputs a policy over which neighboring node should be the next waypoint.}
    \label{fig: model}
    \vspace{-0.4cm}
\end{figure*}

\section{Problem Statement}

Let $\mathcal{E}$ be a bounded environment, classified into free areas $\mathcal{E}_f$ and occupied areas $\mathcal{E}_o $, where $\mathcal{E}_f \cup \mathcal{E}_o = \mathcal{E} $ and $\mathcal{E}_f \cap \mathcal{E}_o = \emptyset$.
The robot maintains a map $\mathcal{M}$ (i.e., robot belief) about $\mathcal{E}$, which is classified into free areas $\mathcal{M}_f $, occupied areas $\mathcal{M}_o$, and unknown areas $\mathcal{M}_u $, where $\mathcal{M}_f \cup \mathcal{M}_o \cup \mathcal{M}_u = \mathcal{M}$.
Starting from an empty belief over the environment (i.e., $\mathcal{M} = \mathcal{M}_u$), the robot executes an exploration path $\psi$ and updates its map/belief about the environment using measurements taken along the path from an onboard sensor (here, a LiDAR).
The exploration task completes when $\mathcal{M}_o$ is closed with respect to $\mathcal{M}_f$\footnote{In practice, the map is often considered complete when all frontiers have been removed, which also ensures the set of occupied areas is closed.}, and the objective is to minimize the \textit{cost} $C(\psi)$ (e.g., the path length $L(\psi)$ or makespan $T(\psi)$) of the exploration path.

Considering the completed exploration path as a finite sequence of robot-visited positions $\psi=[p_1, p_2,...,p_n]$, where $p_i \in \mathcal{E}_f$ is the robot position at decision step $i$, the exploration problem is a partially observable Markov decision process (POMDP). This problem can be formulated as a tuple $(S,A,T,R,\Omega,O)$, where $S$ is a set of states, $A$ is a set of actions, $T$ is a set of conditional probabilities between states, $R$ is a set of reward, $\Omega$ is a set of observations, $O$ is a set of conditional probabilities between states and observations
The true state $s_i=(\mathcal{E}_i,\mathcal{M}_i,\psi_{1:i}), s_i \in S$ is not available to the agent during exploration; instead, the robot only has access to observation (i.e., its map/belief) $o_i=(\mathcal{M}_i,\psi_{1:i}), o_i \in \Omega$, depending on $O(o_i|s_i)$.
At each decision step, the robot selects and takes action $a_i\sim\pi(\cdot|o_i)$, which indicates the next waypoint to visit.
Upon reaching that next state $s_{i+1}$ with $T(s_{i+1}|s_{i}, a_i)$, the agent receives a reward $r_i= -C(\psi_{t-1:t})$.
The objective is to find an optimal policy $\pi^*$, which selects an action at each decision step that maximizes the long-term discounted return $\mathbb{E}(\Sigma^{n}_{t=i} r_t)$.

\section{Methodology}

Solving a highly stochastic POMDP such as autonomous exploration is non-trivial, and thus we propose to rely on model-free DRL for its powerful ability to make implicit predictions based on past experiences.
In this work, we rely on an attention-based neural network similar to~\cite{cao2023ariadne,cao2022catnipp} for its proven ability to model multi-scale dependencies between areas.
We further improve the model's performance by allowing the critic to train based on ground truth information, to help it precisely estimate returns and the underlying transition model.
Guided by these improved predictions, our policy network can finally improve its ability to achieve long-term efficiency under partial observations.

We first decrease the complexity of the problem by extracting a dense collision-free graph from the map.
With this collision-free graph as input, we train an attention-based policy network offline, using the soft actor-critic (SAC) with discrete actions~\cite{haarnoja2018soft,christodoulou2019soft}.
Note that under the actor-critic framework, the critic network is only used during training to assist the policy network in estimating the state-action value.
Therefore, we allow the critic network to access true states $s$, instead of partial observations $o$, to estimate the long-term impact of current actions more precisely, which in turn leads to improved overall performances.
Moreover, benefiting from the nature of attention layers~\cite{vaswani2017attention,vinyals2015pointer}, our policy network is able to generalize to arbitrary graphs and arbitrary scale environments.
We further propose an algorithm to rarefy the dense collision-free graph, allowing our DRL-based planner to naturally scale to larger-scale environments without any additional training.


\subsection{Sequential Decision-making on a Graph}

Consider the collision-free graph $G=(V,E)$ extracted from the map $\mathcal{M}$, where $V=(v_1,v_2,..,v_m), \forall v_i=(x_i,y_i) \in \mathcal{M}_f$ is a set of nodes (one of the nodes is located at the robot current position $p_t$) and $E=((v_1,v_2),...,(v_{m-1},v_m))$ is a set of collision-free edges.
In this work, nodes in $V$ uniformly cover free areas $\mathcal{M}_f$.
We find the $k$ nearest neighbors of each node and check for collisions to construct the set of admissible edges $E$ (i.e., edges between nodes that do not cross $\mathcal{M}_o$ or $\mathcal{M}_u$).
During exploration, this collision graph extends incrementally along with the update of free areas.
Our DRL planner only makes decisions on which neighboring node will be the next waypoint (i.e, $\pi(a_i|o_t)=p(v_i|o_t), (p_t,v_i)\in E$), toward which the robot then navigates.
Note that, during training, the planner only makes decisions upon arriving at the previously selected waypoint, while at execution time, the planner can make decisions at a fixed frequency to be more reactive to map updates.


\subsection{Policy Network}

We first extract an \textit{informative graph} $G^*$, extending the collision graph $G$, as the input of our policy network to allow for more efficient learning.
The informative graph $G^*=(V^*,E)$ shares the same collision-free edge set as $G$, and each node $v^*_i=(v_i,u_i,g_i) \in V^*$ has two more properties than nodes in $V$: 1) the node's \textit{utility} (denoted as $u_i$, quantifying observable frontiers within line of sight from location $v_i$) and 2) the \textit{guidepost} (denoted as $g_i$, a binary value defining whether $v_i$ has already been visited previously).
The utility extracts information from the map to avoid the need for unnecessary learning of low-level pattern recognition, allowing the network to focus on modeling long-term dependencies between areas.
The guidepost representing the path $\psi_{1:t}$ executed so far, which we empirically found to help speed up training by encouraging the robot to select unvisited nodes.
The informative graph is normalized and used as input of our policy network, composed of an encoder and a decoder, as shown in Figure~\ref{fig: model}, where the encoder learns to model the dependencies among nodes $V^*$ and the decoder uses these dependencies to finally output the policy $\pi$ over neighboring nodes.

\noindent \textbf{Encoder} A graph structure is naturally aligned with the input format of an attention layer:
\vspace{-0.3cm}
\begin{equation}
\begin{aligned}
& q_{i}=W^Qh^{q}_{i}, \ k_{i}=W^Kh^{k,v}_{i}, \ v_{i}=W^Vh^{k,v}_{i}, \ u_{ij}=\frac{q_{i}^T\cdot k_{j}}{\sqrt{d}}, \\ & w_{ij}=\left\{\begin{array}{cc}
    \frac{e^{u_{ij}}}{\sum_{j=1}^{n}e^{u_{ij}}}, & M_{ij}=0 \\
    0, & M_{ij}=1
\end{array}\right., \ h'_{i}=\sum_{j=1}^{n}w_{ij}v_{j}, 1\leq i \leq m
\end{aligned}
\label{eq:attention}
\vspace{-0.1cm}
\end{equation}

where each $h_i\in\mathbb{R}^{d\times 1}$ is a $d$-dimension feature vector projected from node $v^*_i$, superscripts $q$ and $k,v$ denote the source of the \textit{query}, \textit{key}, and \textit{value} respectively, $W^Q, W^K, W^V \in \mathbb{R}^{d\times d}$ are learnable matrices, and $M$, which serves as an edge \textit{mask}, is a $m\times m$ adjacency matrix built from $E$ ($M_{ij}=0$, if $(i, j) \in E$, else $M_{ij}=1$).
Note that in each attention layer, each node in the informative graph is only allowed to access the features of its neighboring nodes.
To allow the model to learn dependencies between non-neighboring nodes, we stack six attention layers in the encoder, each taking the output of its previous attention layer as input.
The output of the encoder is a group of \textit{node features}, each $h'_i$ modeling dependencies between node $v^*_i$ and all other nodes.

\noindent \textbf{Decoder} Denoting the node feature corresponding to the robot's current position $p_t$ as $h^c$, we first add a feature vector representing the global graph to it, by passing $h^c$ as the query source and all node features $h'_i$ as the key and value source to an attention layer, concatenating its output with $h^c$ and projecting its back to a $d$-dimension feature $h^{*c}$.
We then select its neighboring node features $h^n_i, \forall (p_t, v_i) \in E$ and input them as the key source and $h^{*c}$ as the query source to a pointer layer~\cite{cao2022dan,cao2022catnipp,cao2023ariadne}, which directly outputs attention weights $w$ as the policy $\pi(a_i|o_t)=p(v_i|o_t), (p_t,v_i)\in E$.
By using this pointer network at the end of the decoder, the final policy's dimensions naturally match the number of neighboring nodes.

\begin{figure}[t]
    \centering
    \vspace{+0.1cm}
    \includegraphics[width=0.45\textwidth]{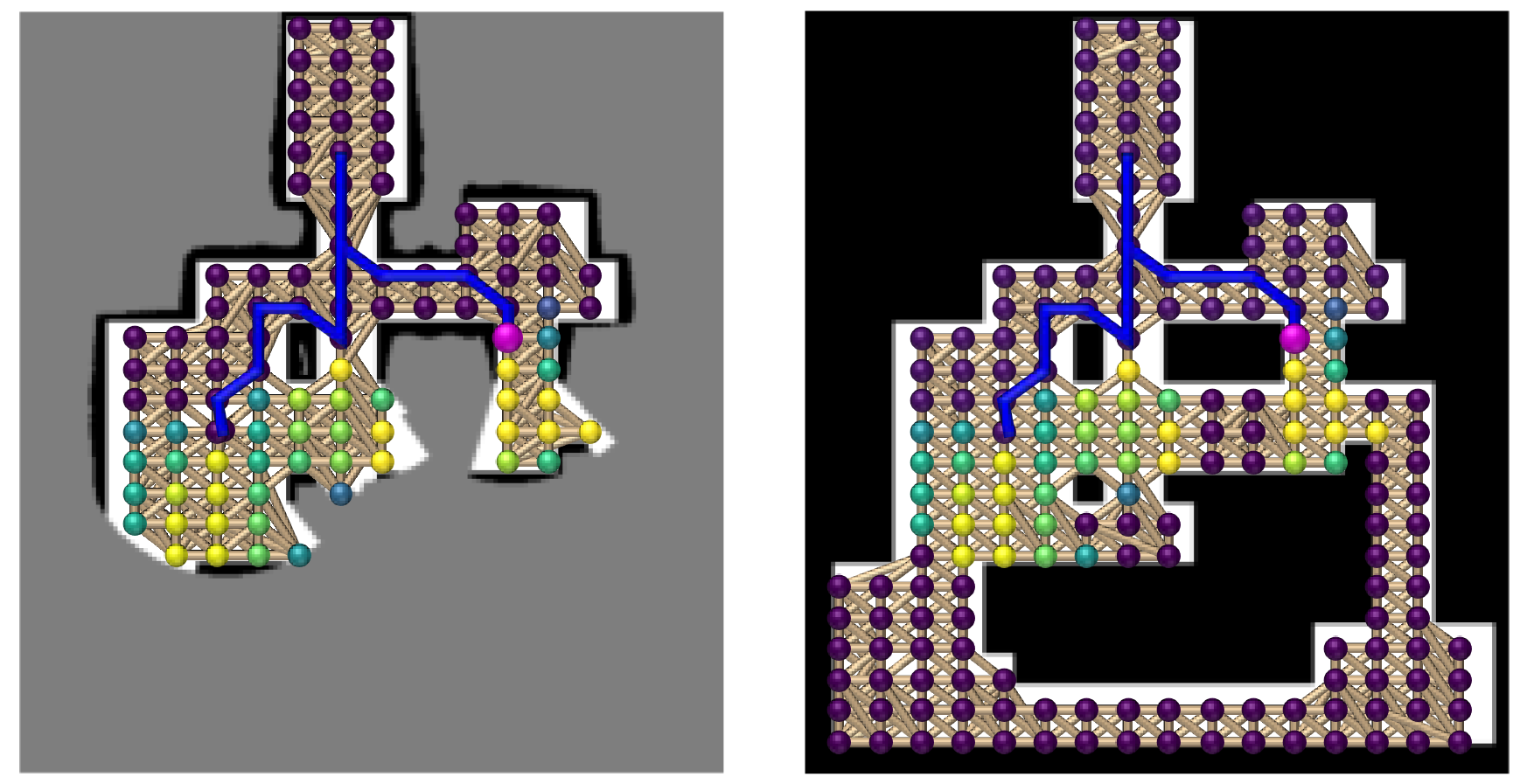}
    \vspace{-0.15cm}
    \caption{
    \textbf{Informative graph (left) and Ground-truth graph (right).} The informative graph is the input of our policy network. The ground truth graph is the input of our critic network, which is only used during training to assist the learning of the policy. Nodes are color-coded based on their utility (dark purple to yellow, low to high).
    The blue trajectory is the path executed so far by the robot (light purple node).}
    \label{fig: network inputs}
    \vspace{-0.4cm}
\end{figure}


\subsection{Training with Ground Truth}

\noindent \textbf{SAC} We train our attention-based policy network using the soft actor-critic (SAC) algorithm with discrete actions~\cite{haarnoja2018soft,christodoulou2019soft}, in the set of small-scale exploration environments provided by~\cite{chen2019self}.
The goal of SAC is to balance the trade-off between maximizing returns and policy entropy:

\vspace{-0.15cm}
\begin{equation}
    \pi^*= \mathop{\rm {argmax}} \mathbb{E}[\sum^T_{t=0}\gamma^t(r_t+\alpha \mathcal{H}(\pi(.|o_t)))],
\vspace{-0.15cm}
\end{equation}

where $\mathcal{H}$ denotes the entropy and $\alpha$ is a temperature parameter that tunes the importance of the entropy term versus the return.
We use reward shaping to help tune the training: for each action $a_t$, in addition to the cost reward $r_c=-C(\psi_{t-1:t})$, we give an exploration reward $r_e$ that quantifies the number of observed frontiers associated with $a_t$.
Upon finishing exploration (i.e., once the utility of all nodes is zero), we also give the agent a fixed finishing reward $r_f$.
Thus the reward used for training is $r_i=a\cdot r_c+b\cdot r_e+r_f$, where $a$ and $b$ are scaling parameters (in practice $a=1/64$, $b=1/50$, $r_f=20$).
SAC trains a critic network $\phi$ to estimate the soft action-state values $Q_{\phi}(o_t, a_i)$, using the critic loss: $J_Q(\phi)=\mathbb{E}_{o_t}[\frac{1}{2}(Q_\phi(o_t,a_t)-(r_t+\gamma \mathbb{E}_{o_{t+1}}[V(o_{t+1})]))^2]$, where $V(o_t) = \mathbb{E}_{a_i}[Q(o_t,a_i)]-\alpha {\rm log}(\pi(\cdot|o_t))$.
The policy network $\theta$ is trained to output a policy that maximizes the expected state-action value, where the policy loss reads: $J_\pi(\theta)=\mathbb{E}_{(o_t,a_i)}[\alpha {\rm log}(\pi_\theta(a_i|o_t))-Q_\phi(o_t,a_i)]$.
The temperature parameter is auto-tuned during training and the temperature loss is calculated as:
$J(\alpha)=\mathbb{E}_{a_t}[-\alpha({\rm log}\pi_t(a_t|o_t)+\overline{\mathcal{H}})]$, where $\overline{\mathcal{H}}$ denotes the target entropy~\cite{haarnoja2018soft,christodoulou2019soft}.

\noindent \textbf{Critic Network} The training of a model-free DRL agent on a POMDP involves the need to estimate state/state-action values from partial observations of the environment, where the model needs to implicitly learn to predict the transition model $T$ and $O$, which is non-trivial in a stochastic problem such as autonomous exploration.
In SAC, the learning of the policy network depends on accurate state-action value estimations from the critic network, where such coupling might further hamper the learning of optimal policies (e.g., from oscillations in thye critic's training from high-variance gradients).
In this context, recent works~\cite{cao2023ariadne} tried to use standard SAC (i.e., identical observation inputs and nearly the same network structure for both actor and critic with just a different final layer to differentiate their output) but found that the critic network remained rather noisy due to the high randomness of the underlying transition model (shown as the gradient variance).
During training, the critic loss remains high ($\sim 10\%$ of $Q(o_t, a_t)$), showing that the critic is struggling at learning the underlying long-term effects of actions/states, and thus providing less accurate estimations to train the actor.

To address this issue, in this work, we leverage the structure of the actor-critic framework, where the critic network is only used to assist policy network training and is not required after being trained, allowing us to take true states as the observations for the critic network (i.e., privileged learning).
Figure~\ref{fig: network inputs} shows a visual comparison between the inputs of the policy network and the critic network.
Specifically, we keep the structure of the vanilla critic network but construct a \textit{ground truth graph} $G'=(V', E')$, from $s_t$ and $o_t$.
$G'$ is similar to the informative graph used as input of the policy network, but the node set $V'$ covers the whole (ground truth) free area in $\mathcal{E}$, the property of each node $v'_i=(x_i, y_i, u_i, e_i)$ is slightly different: the unexplored feature $e_i$ is a binary value indicating whether $v'_i$ is in the already explored area by the robot.
If a node $v'_i$ lies in unexplored areas, we set its utility $u_i=-1$.
By doing so, the critic network now only needs to tackle a more traditional MDP problem, i.e., estimating paths to cover an environment with fully known prior, allowing for more accurate estimations of state-action values (around 10 times lower state-action value loss and gradient variance).
In practice, this translates into $\sim 10\%$ increased average per-step rewards compared to standard SAC (more details can be found in the supplemental material).

\begin{table*}[t]
\vspace{+0.15cm}
\caption{
\textbf{Comparisons with baseline planners in small-scale environments (identical 100 scenarios for each method).}}
\label{table:1}
\vspace{-0.2cm}
\begin{center}
\begin{tabular}{c|c|c|c|c|c|c|c}
\toprule
& Nearest & Utility & NBVP & TARE Local & CNN & ARiANDE & Ours\\
\midrule
Distance (pixel) & 1354($\pm$410) & 1268($\pm$396) & 1323($\pm$371) & 1266($\pm$388) & 1323($\pm$428) & 1204($\pm$378) & \textbf{1118}($\pm321$) \\
\bottomrule
\end{tabular}
\end{center}
\vspace{-0.5cm}
\end{table*}

\noindent \textbf{Training details} Our model is trained on the dungeon environments provided by~\cite{chen2019self}, where each environment $\mathcal{E}$ is a $640\times480$ grid world.
The sensor range of our robot is set to $80$ cells.
We uniformly place $30\times30$ points in the environment and select all points in the explored free area $\mathcal{M}_f$ to form the node set $V$ and construct the information graph $G^*=(V^*, E^*)$, where the number of neighboring nodes $k=20$.
The exploration task is completed once there is no non-zero utility node in $V^*$ (using a threshold to ignore nodes with $u_i\leq 5$ in practice).
During training, we set the max episode length to $128$ decision steps, the discount factor to $\gamma=1$, the batch size to $64$, and the episode buffer size to $2500$.
Training starts after the episode buffer collects more than $1000$ step data.
The target entropy is set to $0.01\cdot{\rm log}(k)$.
Training happens at the end of each episode, and is composed of $8$ iterations.
We use the Adam optimizer with a learning rate of $10^{-5}$ for both policy and critic networks and $10^{-4}$ for the temperature auto-tuning.
The target critic network updates every $64$ training step.
Our model is trained on a desktop equipped with an AMD Ryzen7 5700X CPU and an NVIDIA GeForce RTX 4080 GPU.
We train our model utilizing Ray, a distributed framework for machine learning~\cite{moritz2018ray}, and run 16 training environments simultaneously to accelerate data collection.
The training needs approximately 3 days to fully converge.
Our full code can be found at \url{https://github.com/marmotlab/large-scale-DRL-exploration}.

\subsection{Graph Rarefaction for Large-scale Exploration}

Although our policy network is trained in small-scale environments only, our networks can naturally handle arbitrary graphs.
However, the number of nodes needed to let $G*$ uniformly cover larger-scale environments will be significantly higher, even though while the graph structure would be dense, information would remain sparse (i.e., many frontiers are very far from each other, and separated by a large number of 0-utility nodes).
To extend the use of our trained model to larger-scale environments, we propose a method to rarefy our informative graph, which lets the model capture longer-term dependencies more efficiently (i.e., frontiers far away can connect with each other through fewer edges).
In short, we first classify all nodes with non-zero utility into multi groups according to neighboring relationship, and then find the shortest paths on $G$ from the robot's position to each group through A* and check for line of sight at each waypoint in these paths, to find a minimal set of nodes $V^s \subset V^*$ that represents all information in the map (the pseudo-code can be found in the supplemental material).
The result is a sparser information graph $G^s=(V^s, E^s)$, which is used as the policy network's input. 

\section{Experiment}

\begin{figure*}[t]
    \centering
    \subfloat[DSVP, distance $1351m$, time $796s$]{\includegraphics[width=0.32\textwidth]{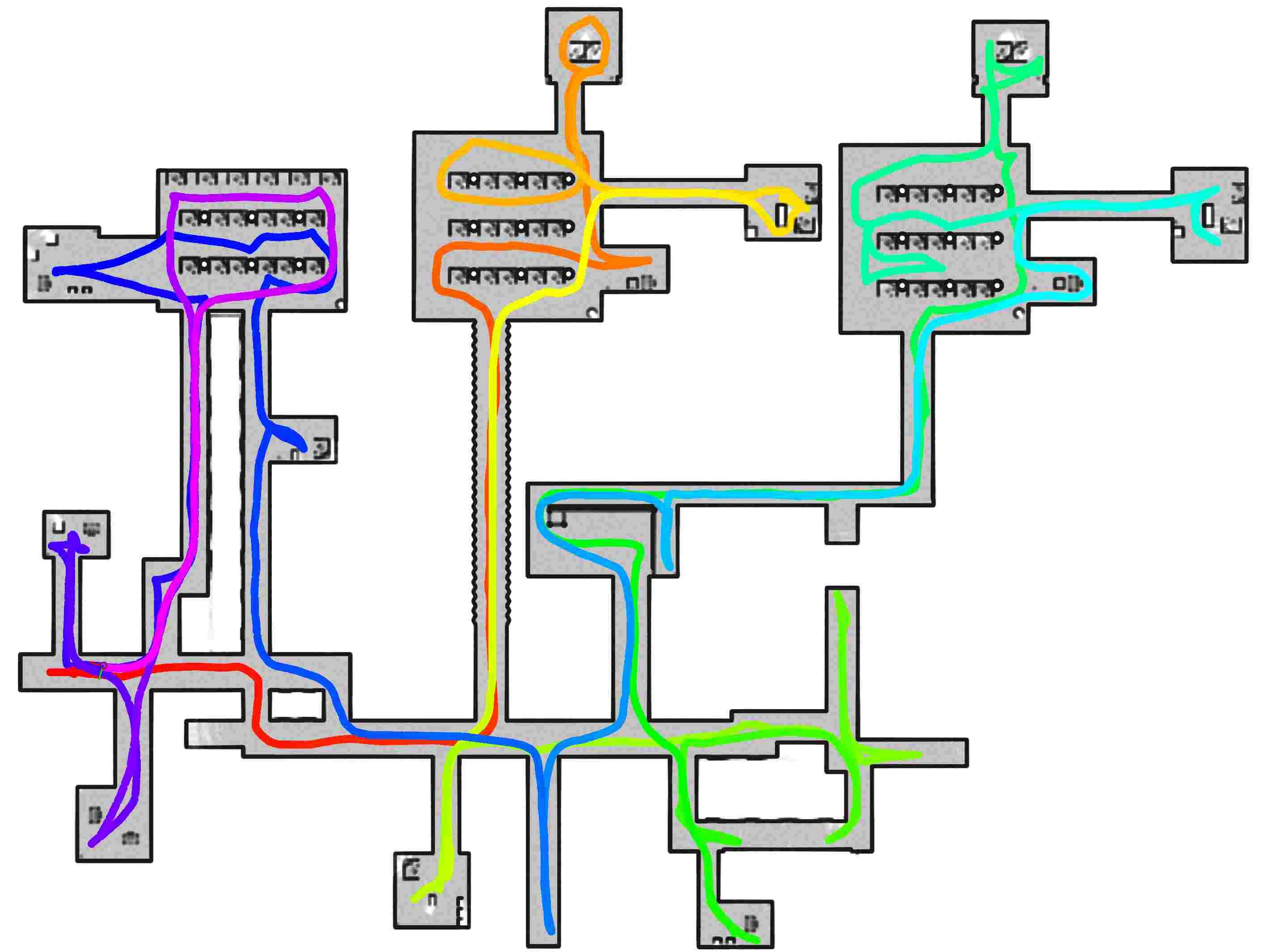}
    }
    \hfill
    \subfloat[TARE, distance $1177m$, time $644s$]{\includegraphics[width=0.32\textwidth]{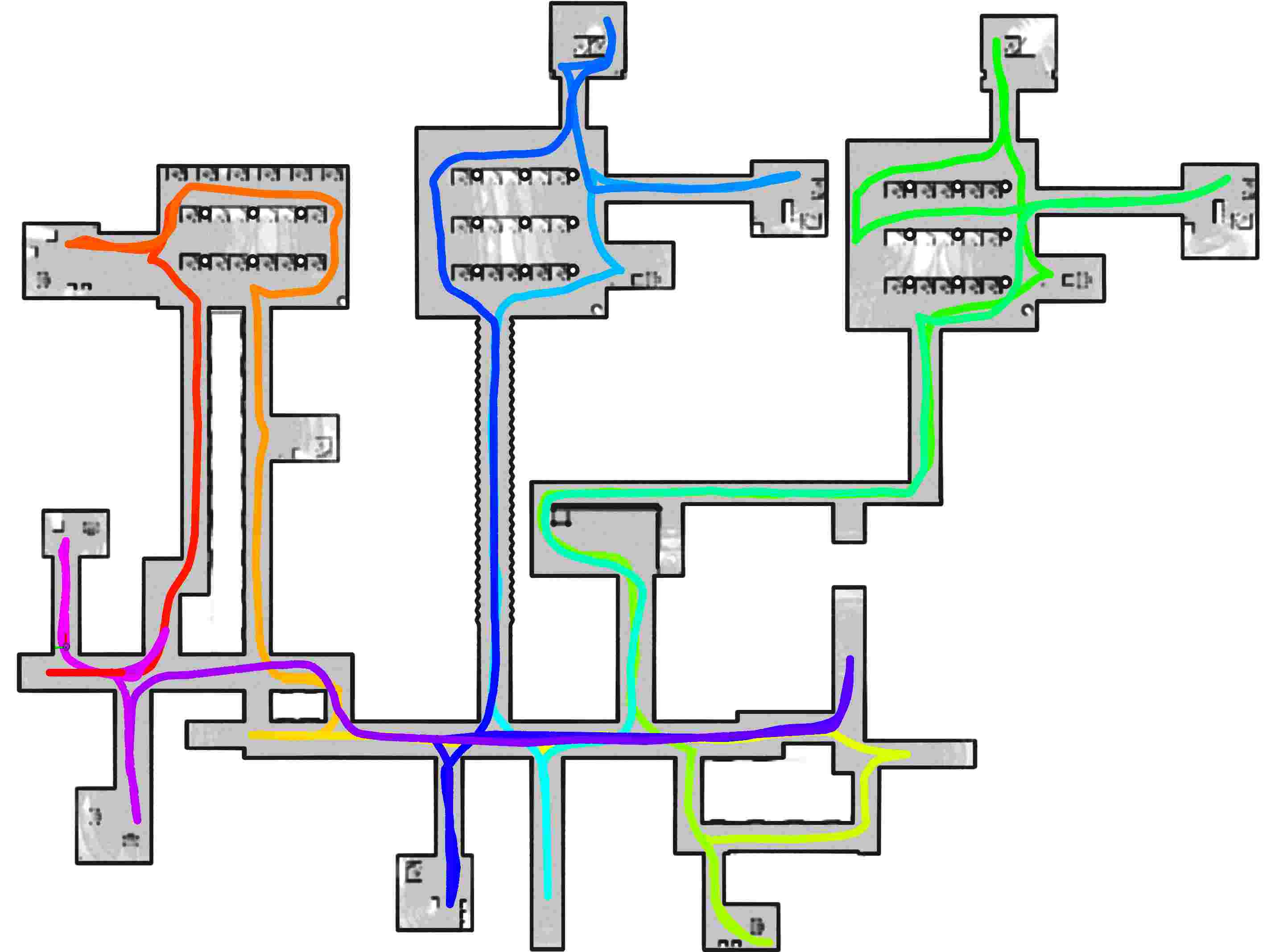}}
    \hfill
    \subfloat[Ours, distance $985m$, time $582s$]{\includegraphics[width=0.32\textwidth]{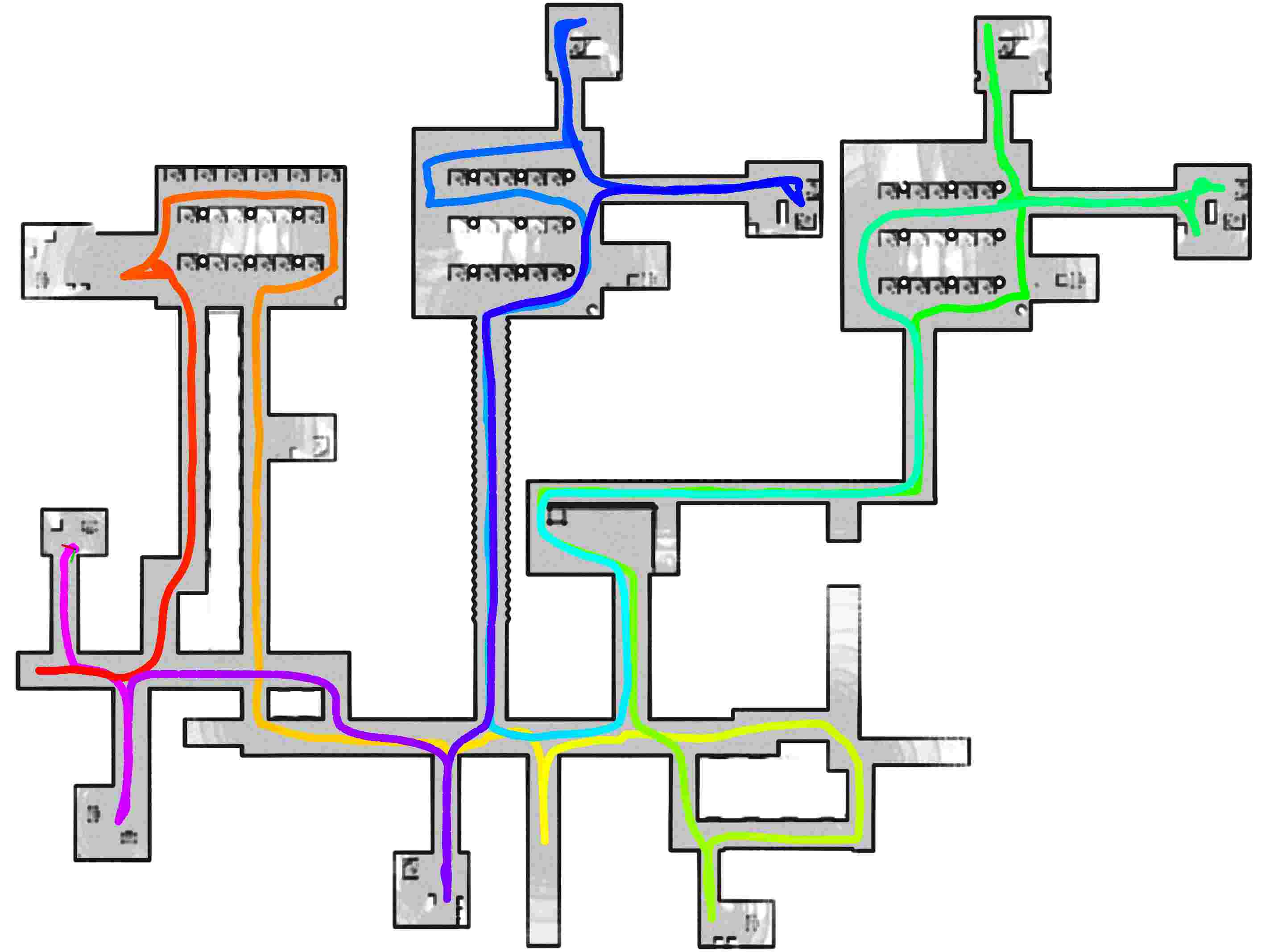}}
    \caption{
    \vspace{-0.2cm}
    \textbf{Exploration paths comparisons in a large-scale $130m\times 100m$ indoor office simulation.}}
    \label{fig: trajectory comparison}
    \vspace{-0.6cm}
\end{figure*}

\begin{figure}[t]
    \centering
    \subfloat[TARE, $1505m$, $803s$]{\includegraphics[width=0.225\textwidth]{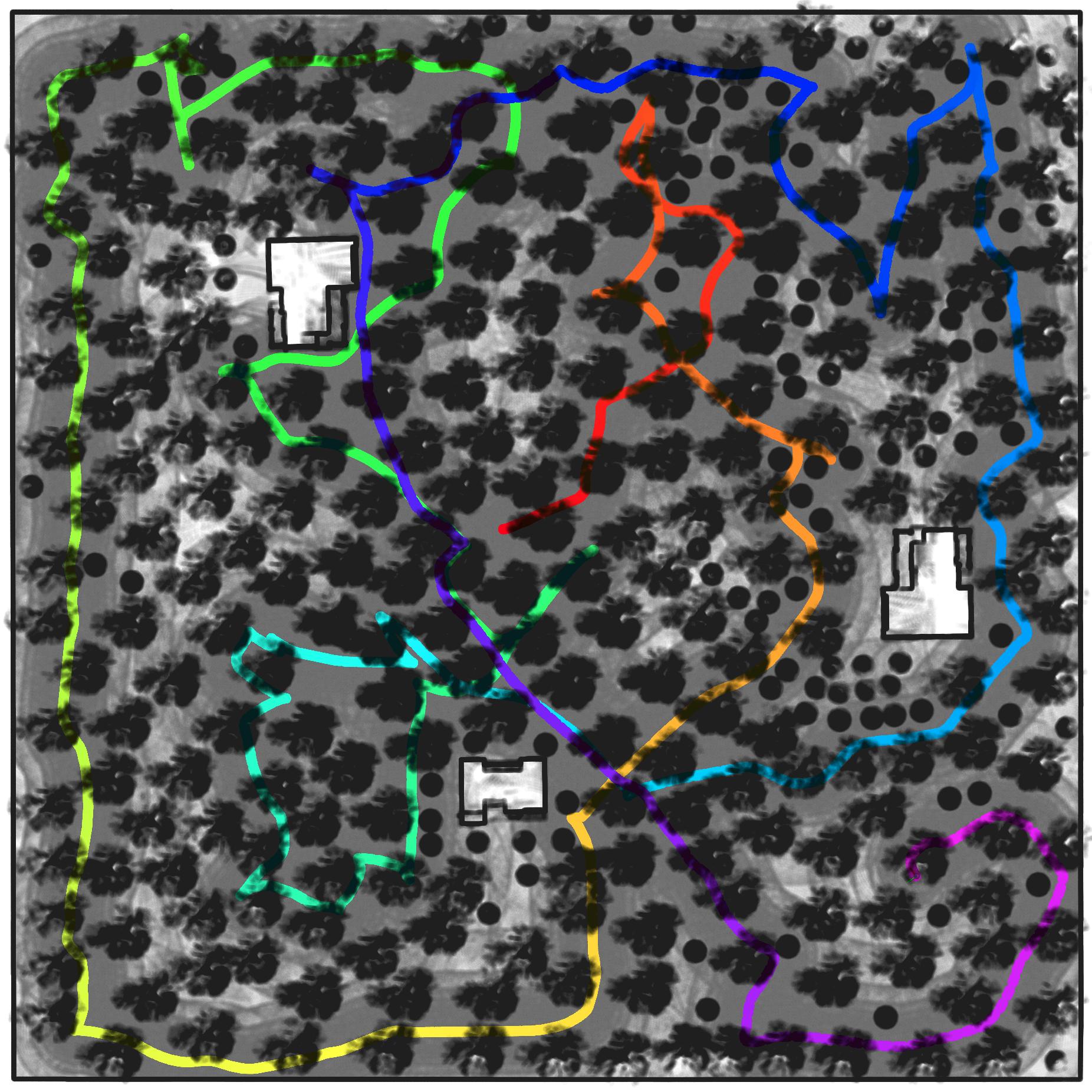}}
    \hfill
    \subfloat[Ours, $1104m$, $659s$]{\includegraphics[width=0.228\textwidth]{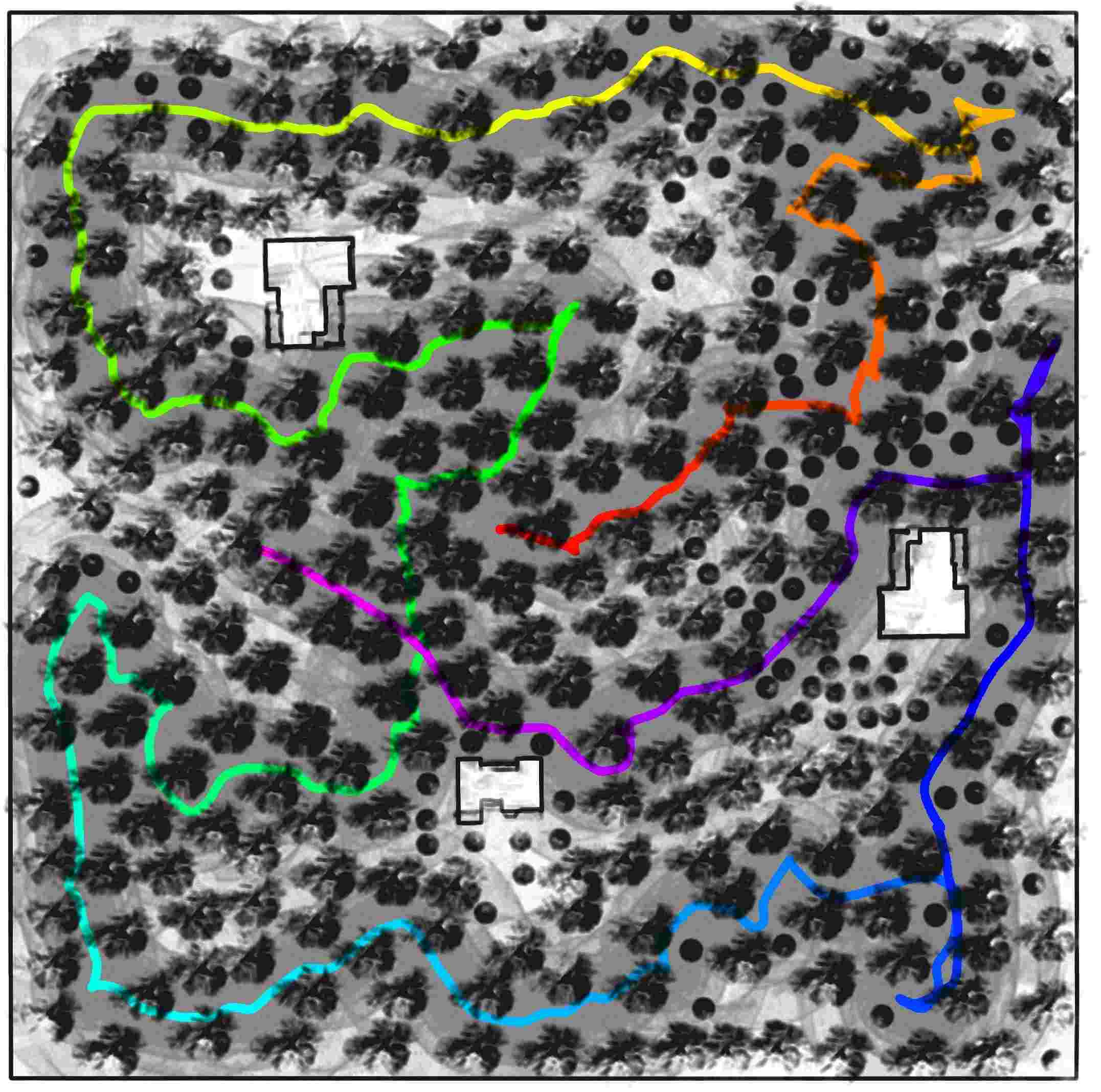}}
    \caption{
    \textbf{Exploration paths comparisons in a large-scale $150m\times 150m$ outdoor forest simulation.}}
    \label{fig: outdoor}
    \vspace{-0.5cm}
\end{figure}

\subsection{Validation in Small-scale Environments}

We first test our trained model on a benchmark set of $100$ small-scale simplified simulation environments (never seen during training).
We compare our planner with baseline planners including (1) \textit{nearest}: the robot always moves to the nearest frontier, (2) \textit{utility}: the robot moves to the frontier which balances cost and utility, (3) \textit{NBVP}: a sampling-based planner based on rapid random trees~\cite{bircher2016receding}, and (4) \textit{TARE Local}: the local level planner of TARE~\cite{cao2021tare,cao2023representation}. (5) CNN: a DRL planner based on CNNs~\cite{chen2019self}. (6) ARiADNE: an ablation variant of our model trained without ground truth as in~\cite{cao2023ariadne}).

In our tests, we consider exploring more than 99\% (meaning minor error is tolerable) of the free area in the environment as completing the exploration.
As shown in Table~\ref{table:1}, our DRL planner outperforms all baseline planners (11\% better than TARE Local and Utility, 15\% better than CNN and NBVP) in terms of average travel distance to complete the exploration.
We first note that the vanilla model trained without ground truth already achieves shorter exploration paths than all other baseline planners, showing the efficiency of our attention-based policy network to reason about dependencies of areas at multiple scales and thus making decisions that balance exploration and refining the map.
We then note that our model trained with ground truth further improves over this vanilla ARiADNE model (7\% better), verifying that training the critic network with ground truth can assist the policy network to estimate the transition model and long-term returns more precisely.
Our results suggest that our planner can effectively predict the structure of unexplored areas, to make further non-myopic decisions and avoid redundant movements, by relying on past experiences in similarly structured indoor environments seen at training.
We visualize attention weights of our attention layer to observe the insight learned mechanism (see our supplemental material), and we find that in some heads the planner focuses its attention over areas where connections might be missing, implying that the planner is predicting the structure of unknown areas from the belief of known areas.

\subsection{Validation in Large-scale Environments}

\begin{figure*}[t]
    \vspace{+0.1cm}
    \centering
    \includegraphics[width=\textwidth]{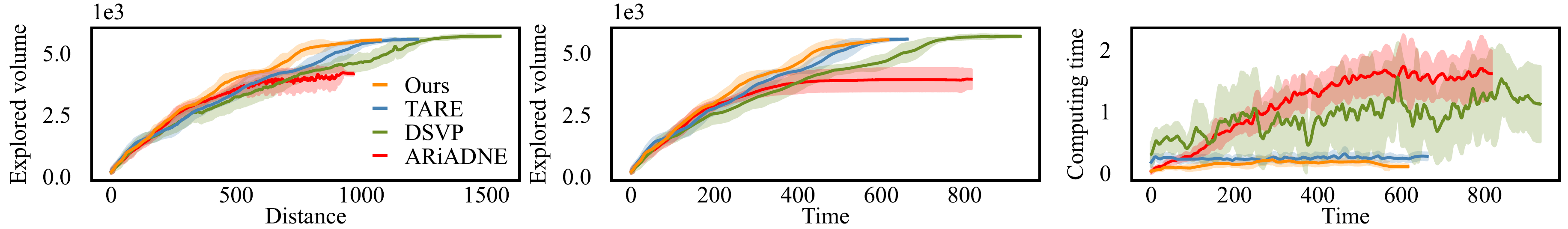}
    \vspace{-0.6cm}
    \caption{
    \textbf{Comparisons with baseline planners in the large-scale indoor benchmark~\cite{cao2023representation} (10 runs each), including ARiADNE despite its inability to complete exploration.}}
    \label{fig: comparison with large scale planners}
    \vspace{-0.3cm}
\end{figure*}

We then test our large-scale exploration planner in a Gazebo simulation of a $130m\times 100m$ office with cluttered obstacles provided by~\cite{cao2023representation}.
Compared to our small-scale test environments where the sensor model and robot kinetic are highly simplified, in this large-scale test environment, the simulation considers real-world sensor model (a Velodyne 16-line 3D LiDAR) and real robot constraint (a four-wheel differential drive robot with max speed $2m/s$).
Note that the model used for our large-scale tests is the same as the one for our small-scale tests, but using our graph rarefaction algorithm.
We use Octomap~\cite{hornung13auro} to classify free and occupied areas (i.e., define ''explored'' area).
We set the resolution of Octomap to $0.4m$, the max map update range to $20m$, the gap of nodes to $2m$, the number of neighboring nodes in the dense collision-free graph to $5$, the number of neighboring nodes in the sparse information graph to $10$, the sparse radius to $12m$, and replan the robot's path every $0.8s$.
Although we construct a 3D Octomap, the planner only considers frontiers at the ground level (i.e., planning is based on a 2D belief).
All computations are conducted on a Intel i7-1270p CPU.

\begin{table}[t]
\caption{
\textbf{Comparisons with baseline planners in the large-scale indoor benchmark~\cite{cao2023representation} (10 runs each).}}
\vspace{-0.2cm}
\label{table:2}
\begin{center}
\begin{tabular}{c|c|c|c|c}
\toprule
& DSVP & TARE & Ours & Human\\
\midrule
Distance ($m$) & 1462 & 1158 & \textbf{1020} & 879 \\
Time ($s$) & 870 & 634 & \textbf{590} & 520 \\
Computing ($s$) & 0.90 & 0.24 & \textbf{0.15} & / \\
Efficiency ($m^3/m$) & 3.91 & 4.82 & \textbf{5.42} & 6.27 \\
Efficiency ($m^3/s$) & 6.57 & 8.81 & \textbf{9.36} & 10.60 \\
\bottomrule
\end{tabular}
\end{center}
\vspace{-0.6cm}
\end{table}

\begin{table}[t]
\caption{
\textbf{Comparisons with baseline planners in the large-scale outdoor benchmark~\cite{cao2023representation} (5 runs each).}}
\label{table:3}
\vspace{-0.3cm}
\begin{center}
\begin{tabular}{c|c|c|c|c}
\toprule
& DSVP & TARE & Ours & Human\\
\midrule
Distance ($m$) & 2002 & 1347 & \textbf{1174} & 1071 \\
Time ($s$) & 1053 & 707 & \textbf{686} & 549\\
Computing ($s$) & 1.31 & 0.70 & \textbf{0.44} & / \\
Efficiency ($m^3/m$) & 21.13 & 29.37 & \textbf{32.63} & 36.05 \\
Efficiency ($m^3/s$) & 40.17 & \textbf{55.96} & 55.85 & 70.33 \\
\bottomrule
\end{tabular}
\end{center}
\vspace{-0.8cm}
\end{table}

\begin{figure*}[t]
\centering
    \subfloat[Turtlebot3]{\includegraphics[width=0.165\textwidth]{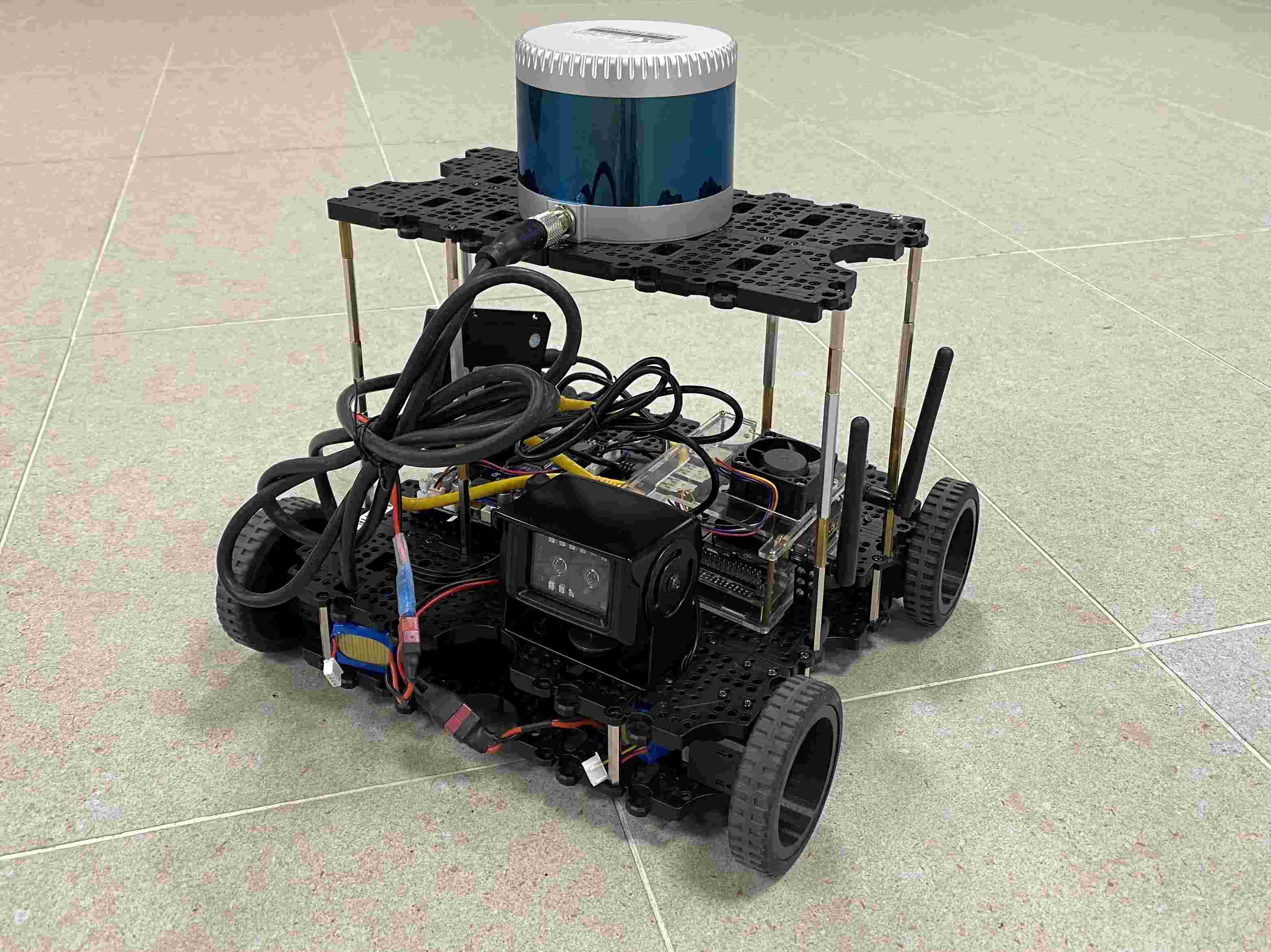}}
    \hfill
    \subfloat[Teaching room]{\includegraphics[width=0.165\textwidth]{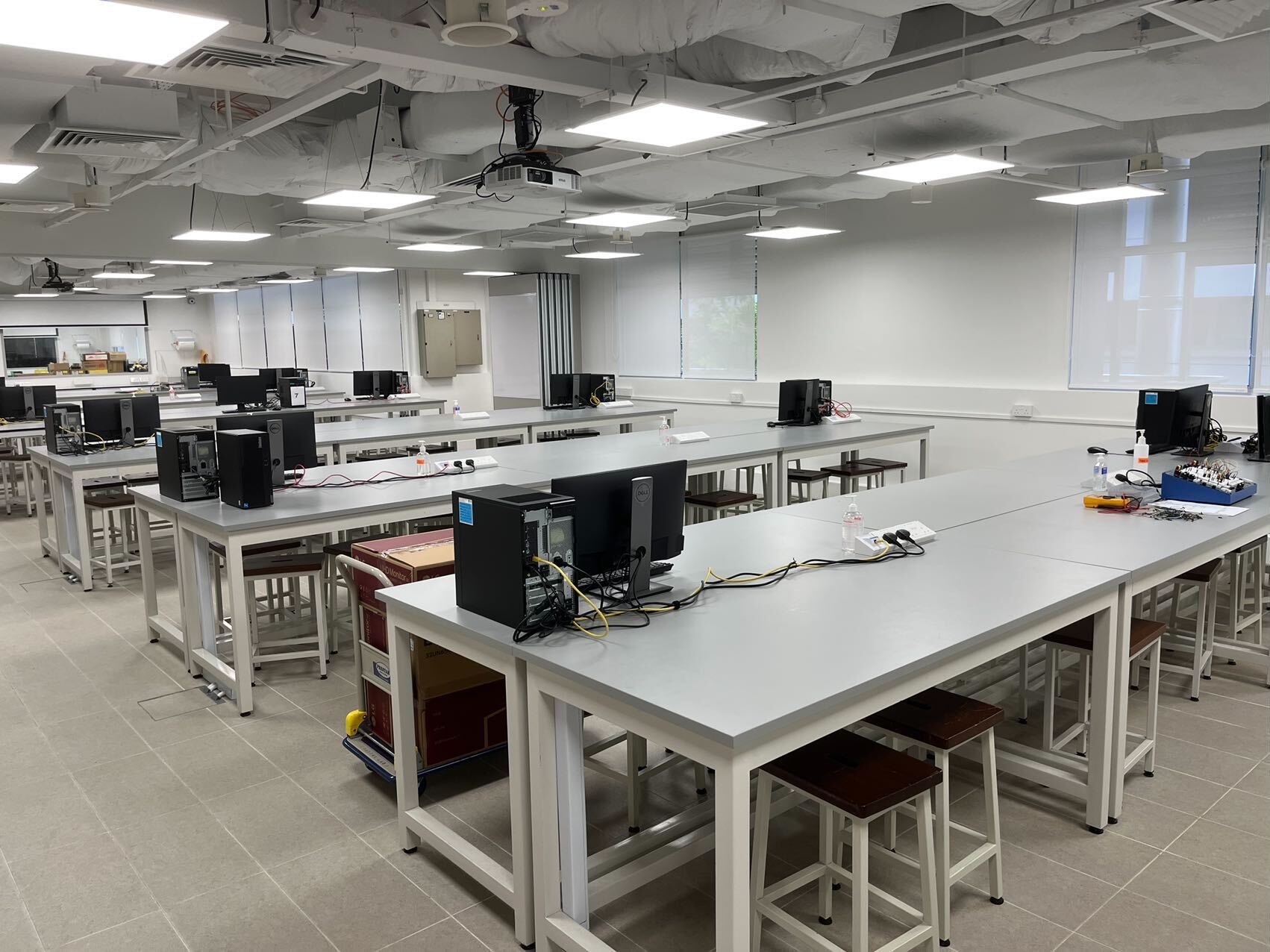}}
    \hfill
    \subfloat[Final Octomap] {\includegraphics[width=0.6\textwidth]{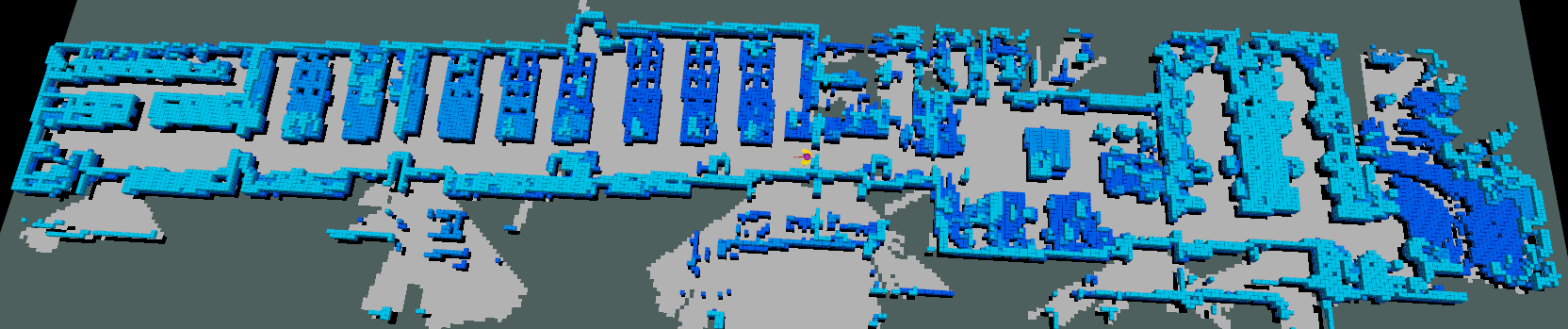}} \\
    \vspace{-0.2cm}
    \subfloat[Office room]{\includegraphics[width=0.165\textwidth]{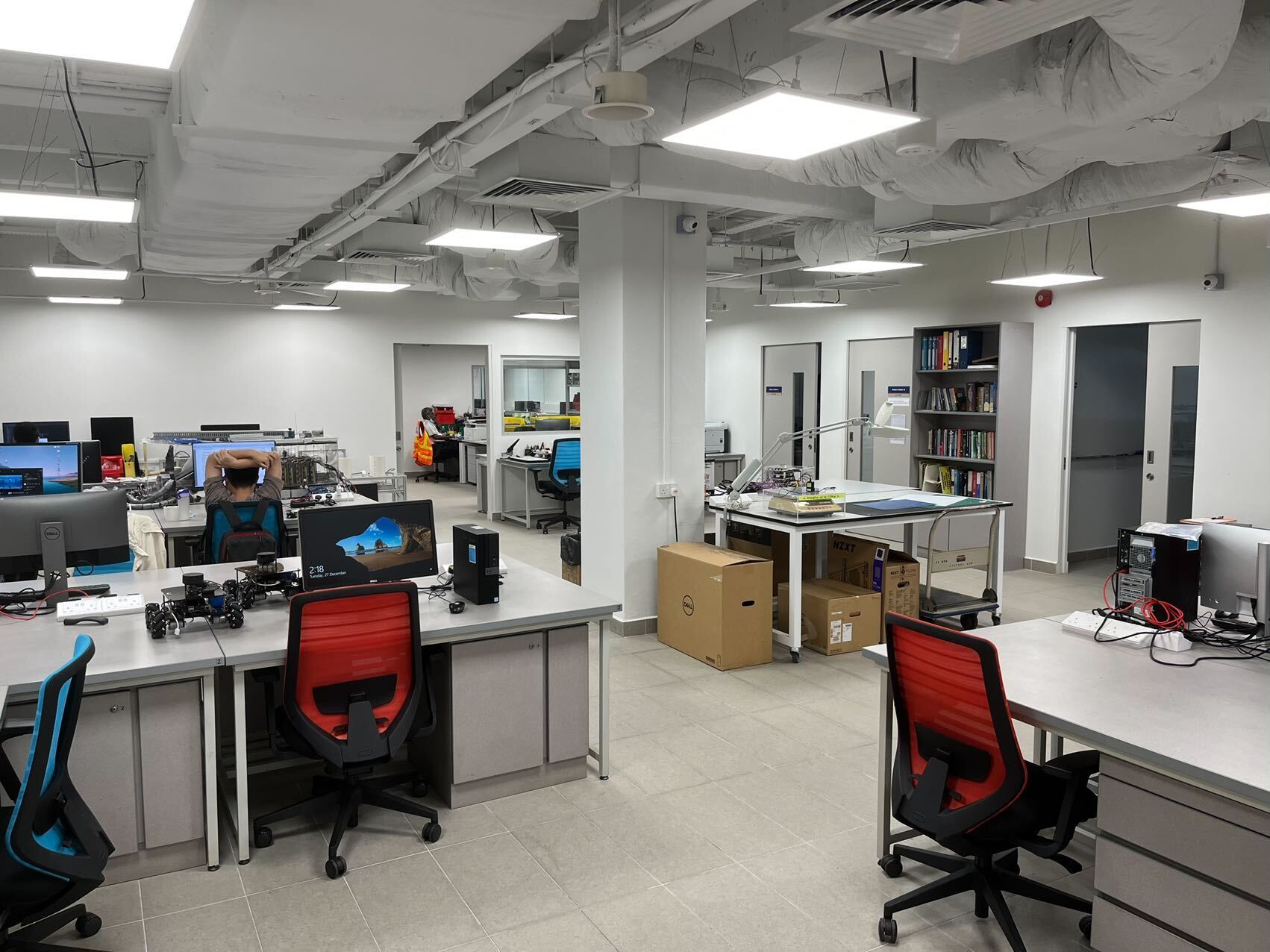}}
    \hfill
    \subfloat[Another office room]{\includegraphics[width=0.165 \textwidth]{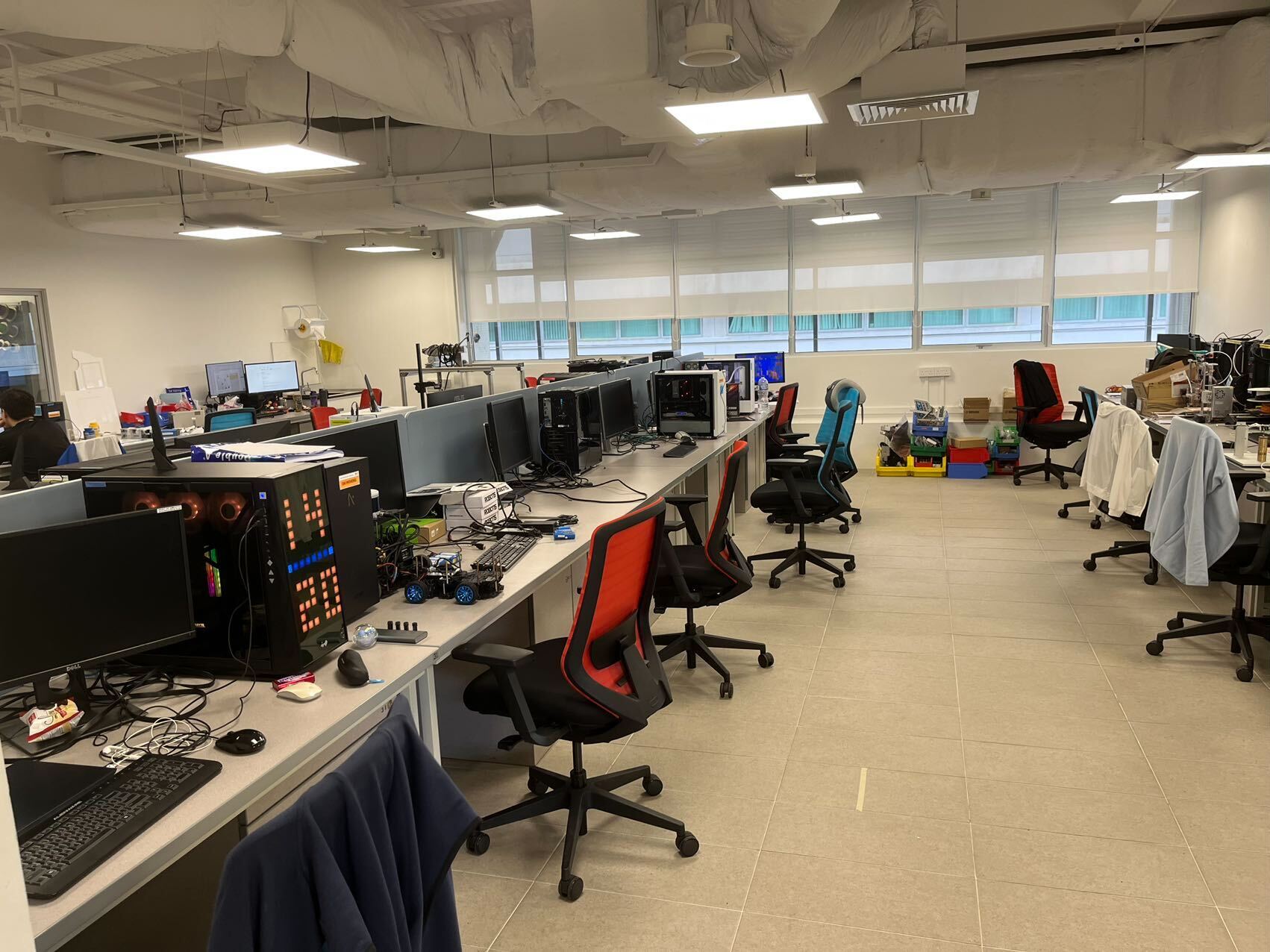}}
    \hfill
    \subfloat[Final point cloud map]{\includegraphics[width=0.6\textwidth]{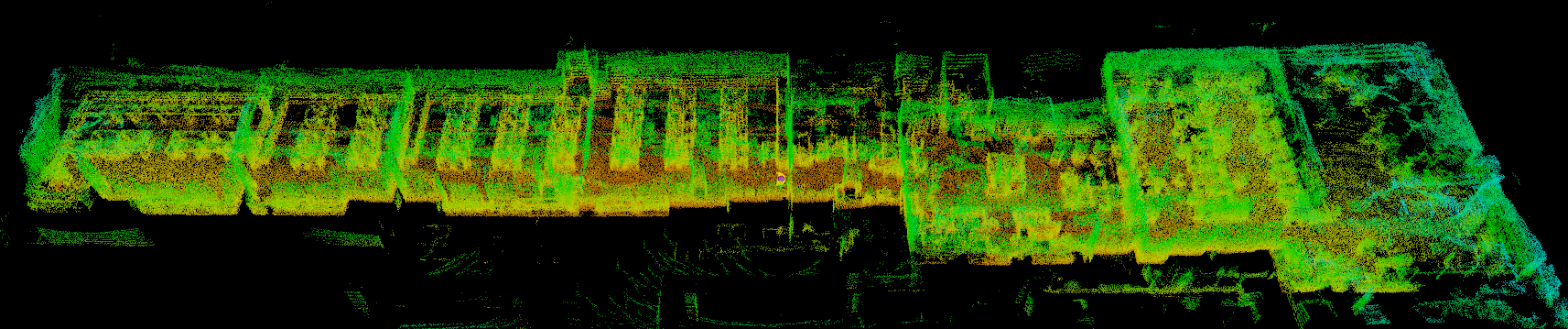}}
    \caption{
    \textbf{Real world experiment in a $80m\times 10m$ laboratory with cluttered furniture and moving pedestrians.}}
    \vspace{-0.4cm}
    \label{fig: real-world experiment}
\end{figure*}

We compare our large-exploration planner with: (1) TARE~\cite{cao2023representation,cao2021tare} (also the provider of this test environment), the state-of-the-art large-scale exploration planner which in general produces $80\%$ increased exploration efficiency over benchmark planners such as~\cite{bircher2016receding,dang2020graph}; in TARE, a global planner is designed to find a global TSP path to visit all unexplored areas, and the start of the global path is taken as the destination for the local path planner discussed in the last subsection, (2) DSVP~\cite{zhu2021dsvp}, which uses a rapid random tree-based local planner and a graph-based global planner, (3) Best human practice with full prior knowledge about the environment provided by~\cite{cao2023representation} as the reference for optimal exploration path.
The hyperparameters of TARE and DSVP were tuned by the original authors directly on benchmark environments. Note that TARE further considers 3D frontiers but still does path planning in 2D action space.
We evaluate the performance of all planners from multiple metrics: (1) \textit{Distance}: total exploration path length, (2) \textit{Time}: total duration of the exploration task, (3) \textit{Computing}: per-step planning time, and (4) \textit{Efficiency}: explored volume divided by the travel distance or makespan.
Results are shown in Table~\ref{table:2}, Figure~\ref{fig: trajectory comparison}, and Figure~\ref{fig: comparison with large scale planners}.

We note that our DRL-based planner outperforms TARE in terms of exploration efficiency (12\% in terms of distance efficiency and 6\% in terms of time efficiency) and algorithm computing time (including frontier detection, informative graph update, and network inference time, 60\% faster even though our planner is implemented in Python and TARE in C++).
As shown in Figure~\ref{fig: trajectory comparison}, the exploration path planned by our DRL planner exhibits fewer redundant movements than benchmark planners.
Although TARE theoretically guarantees near-optimal paths to cover frontiers in its current belief, we believe the advanced performance of our DRL planners shows the key importance of predicting the potential structure of unknown areas, which can yield decisions that benefit longer-term exploration.
Note that paths planned by TARE allow the robot to move at the fastest speed, as TARE explicitly considers the motion constraint of the robot.
Therefore, our advantage in terms of time efficiency is smaller than on distance efficiency.

Although our model was trained in indoor environments, we also test it in a $150m\times 150m$ cluttered outdoor forest Gazebo environment without further training (results are shown in Table~\ref{table:3} and Figure~\ref{fig: outdoor}).
In this outdoor environment, the structure learned by our model (e.g., junctions and corridors) in indoor environments does not exist anymore, and we expect our planner not to be able to perform implicit predictions as well as in indoor environments.
As expected, we observe our planner sometimes randomly chooses waypoints, implying that it cannot decide which action is optimal. However, in this cluttered outdoor environment, our DRL planner still achieves time efficiency on par with TARE and better distance efficiency. 

\vspace{-0.4cm}
\subsection{Validation in Real World}

We finally validate our large-scale planner in the real world, using a wheeled robot equipped with a 3D LiDAR to explore an $80m\times 10m$ indoor laboratory with cluttered furniture and moving pedestrians (see Figure~\ref{fig: real-world experiment}).
The robot platform is a customized four-wheel Turtlebot3 with a max speed of $0.2m/s$.
We use a Leishen C16 LiDAR with LOAM~\cite{zhang2014loam} to both get LiDAR odometry and mapping.
For our DRL planner, compared to the parameters used in simulation, we set the resolution of Octomap to $0.2m$, and the gap of nodes to $0.8m$ to better handle such a cluttered real-world indoor environment.
We believe the robot successfully and efficiently explores the lab in $12$ minutes, highlighting the robust applicability of our planner in the real world.
It should be noted that most previous DRL-based planners~\cite{zhu2018deep,li2019deep,chen2019self,chen2020autonomous,xu2022explore} are not validated on hardware in such a cluttered and large-scale environment. Compared to these planners that directly take the map as network input, where images from the real world may be very noisy and different from those used during training, we believe that extracting an informative graph helps the learned model avoid failure on never-seen real-world scenarios.

\section{Conclusion}

In this work, we propose a DRL-based exploration framework, which relies on ground truth information to assist the training of an attention-based policy network.
We show that our trained model can more precisely estimate the unexplored areas, which helps the planner better reason about its map/belief and make decisions that benefit long-term objectives.
With a graph-rarefaction algorithm, our planner becomes the first DRL-based method applicable to large environments and significantly outperforms state-of-the-art conventional planners in a benchmark indoor simulation environment.
We also validate our planner in the real world, highlighting its potential for real-life deployments on robot.

Future works will focus on extending our planner from single to multi-robot exploration, where the planner needs to further help robots achieve efficient cooperation.
We are also interested in letting the robot take account of the robot's motion constraint in training.  

\section{Acknowledgement}
This work was supported by Temasek Laboratories (TL@NUS) under grant TL/FS/2022/01 and the Singapore Ministry of Education Academic Research Fund Tier 1.

\appendix

\begin{figure}[h]
    \centering  
    \includegraphics{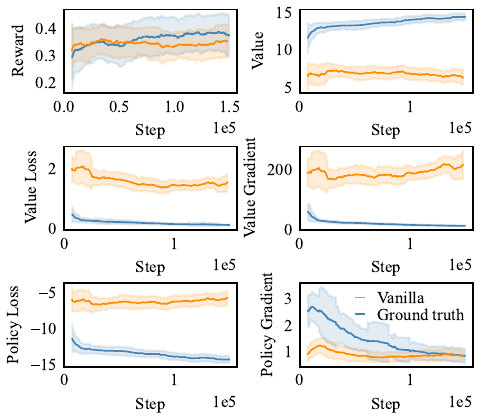}
    \caption{
    \textbf{Ablation results, with and without ground truth when training the critic network.}
    By simplifying the POMDP to a MDP, our privileged learning approach reduces the state-action value loss and gradient variance by 10 times over the original critic network.}
    \label{fig: ablation}
\end{figure}

\begin{figure}[h]
    \centering   \includegraphics[width=0.48\textwidth]{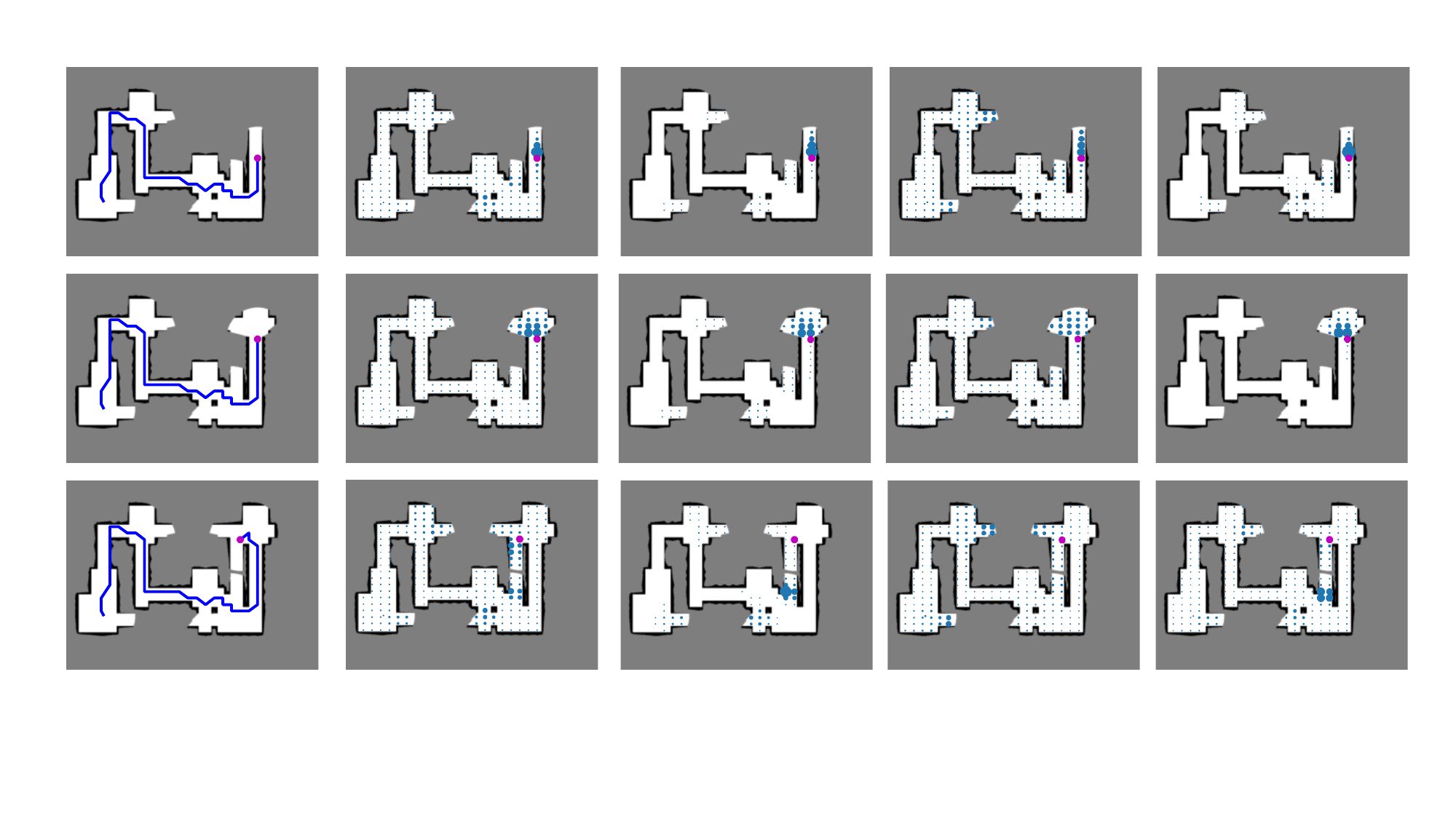}
    \caption{
    \textbf{Visualization of the learned attention weights from the three main heads of our decoder in a small-scale scenario.}
    The query source is the node feature corresponding to the current position (purple).
    Key and value sources are all nodes in the graph (blue), and their radius is proportional to their learned attention weight.
    Note how the first and second heads adapt their attention when they notice that two corridors (top right) might connect.}
    \label{fig: attention visualization}
\end{figure}

\begin{algorithm}[h]
\caption{Dense Graph Rarefaction Algorithm}
\label{alg: building sparse graph}
\KwIn {node set $V^*$, map $\mathcal{M}$, robot pos. $p_t$, radius $D_{th}$}
\KwOut {sparse information graph $G^s=(V^s,E^s)$}
Get non-zero utility node set $U$ from $V^*$ \\
Initialize node set $V^s \gets U$, covered node set $\overline{U} \gets \emptyset$\\
\For{$v \in U$}{
    \lIf {$v \in \overline{U}$}{continue}
    Find nearby node set $N$ in $D_{th}$ \\
    \For {$v' \in N$}{
        \lIf {line $(v,v')$ is collision free}{
            $\overline{U} \gets v'$}}
    Find path $\zeta$ from $p_t$ to $v^*$, set ref node $v_{ref}=v$\\
    \For {$i \in |\zeta|$}{
        \uIf {line $(v_{ref},\zeta_i)$ is not collision free or $L(v_{ref},\zeta_i) \geq D_{th}$}{
            $v_{ref}=\zeta_{i-1}$, $V^s \gets v_{ref}$}
    }
}
Get collision free edge set $E^s$ based on $V^s$ and $\mathcal{M}$
\end{algorithm}
 
\bibliographystyle{IEEEtran}
\bibliography{references}

\end{document}